\documentclass[runningheads]{llncs}
\usepackage[T1]{fontenc}
\usepackage{times}
\usepackage{epsfig}
\usepackage{graphicx}
\usepackage{amsmath}
\usepackage{amssymb}
\usepackage{multirow}
\usepackage{booktabs}
\usepackage{makecell}
\usepackage{diagbox}
\usepackage[pagebackref=true,breaklinks=true,letterpaper=true,colorlinks,bookmarks=false]{hyperref}
\begin{document}
\title{${M}^{3}$:Manipulation Mask Manufacturer for Arbitrary-Scale Super-Resolution Mask}
\titlerunning{${M}^{3}$}
%
\author{Xinyu Yang\inst{1}\and
Xiaochen Ma\inst{1} \and
Xuekang Zhu\inst{1} \and
Bo Du\inst{1} \and
Lei Su\inst{1}\and
Bingkui Tong\inst{1}\and
Zeyu Lei\inst{1,2}\and
Jizhe Zhou\inst{1,3}\thanks{Corresponding author.}}
\authorrunning{X. Yang et al.}

\institute{College of Computer Science, Sichuan University, China \email{2021141460237@stu.scu.edu.cn} \and
Department of Computer and Information Science, University of Macao, Macao SAR \and
Engineering Research Center of Machine Learning and Industry Intelligence, MOE, China
\email{jzzhou@scu.edu.cn}\\}

\maketitle              
\begin{abstract}
In the field of image manipulation localization (IML), the small quantity and poor quality of existing datasets have always been major issues. A dataset containing various types of manipulations will greatly help improve the accuracy of IML models. Images found on public forums, such as those in online image modification communities, are often manipulated using various techniques. Creating a dataset from these images can significantly enhance the diversity of manipulation types in our data. However, due to resolution and clarity issues, images obtained from the internet often contain noises, making it difficult to obtain clean masks by simply subtracting the manipulated image from the original. These noises are difficult to remove, rendering the masks unusable for IML models. Inspired by the field of change detection, we treat the original and manipulated images as changes over time for the same image and view the data generation task as a change detection task. Due to clarity issues between images, conventional change detection models perform poorly. Therefore, we introduced a super-resolution module and proposed the Manipulation Mask Manufacturer (MMM) framework, which enhances the resolution of both original and tampered images to improve comparison. Simultaneously, the framework converts the original and tampered images into feature embeddings and concatenates them, effectively modeling the context. Additionally, we used our MMM framework to create the Manipulation Mask Manufacturer Dataset (MMMD), which covers a wide range of manipulation techniques. We aim to contribute to the fields of image forensics and manipulation detection by providing more realistic manipulation data through MMM and MMMD. Detailed information about MMMD and the download link can be found at: https://github.com/ndyysheep/MMMD.

\end{abstract}
\section{Introduction}

\begin{figure*}[h]
\centering
\includegraphics[width=0.8\textwidth]{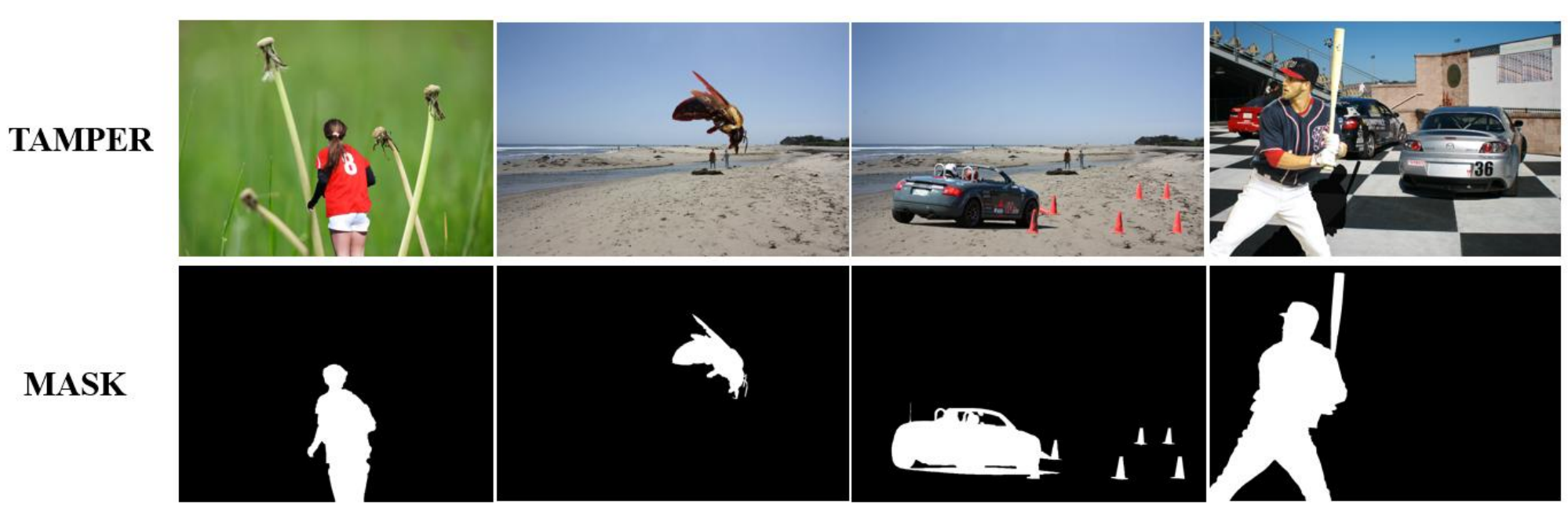}
\caption{Tampered images and their corresponding masks for the image manipulation localization task.} 
\label{IML}
\end{figure*}

Advances in digital image processing~\cite{digital_image_processing} have made software like Adobe Photoshop~\cite{Adobe_Photoshop} and GIMP~\cite{GIMP} more powerful, facilitating widespread image manipulation. Increasingly, there are examples of false information, retouched photographs, or edited video being released on social media. In many cases, this information goes "viral" in just days, even hours~\cite{wild_spread}. This proliferation of false information and manipulated images threatens public knowledge, trust, and safety. Thus, image manipulation localization has emerged, and in some literature, it is also referred to as "forgery detection~\cite{forgery_detection}" or "tamper detection~\cite{tamper_detection}." Its purpose is to discern whether an input image is manipulated or authentic and to depict the exact manipulated parts of an image through a mask~\cite{su}. These parts are semantically different from the original content (the original image before manipulation). It does not include purely generated images (e.g., images generated from pure text) or the introduction of noise or other non-semantic changes through image processing techniques that do not alter the underlying meaning of the image. Standard tampered images and their masks are shown in Fig. \ref{IML}.

\begin{figure*}[t]
\centering
\includegraphics[width=1.0\textwidth]{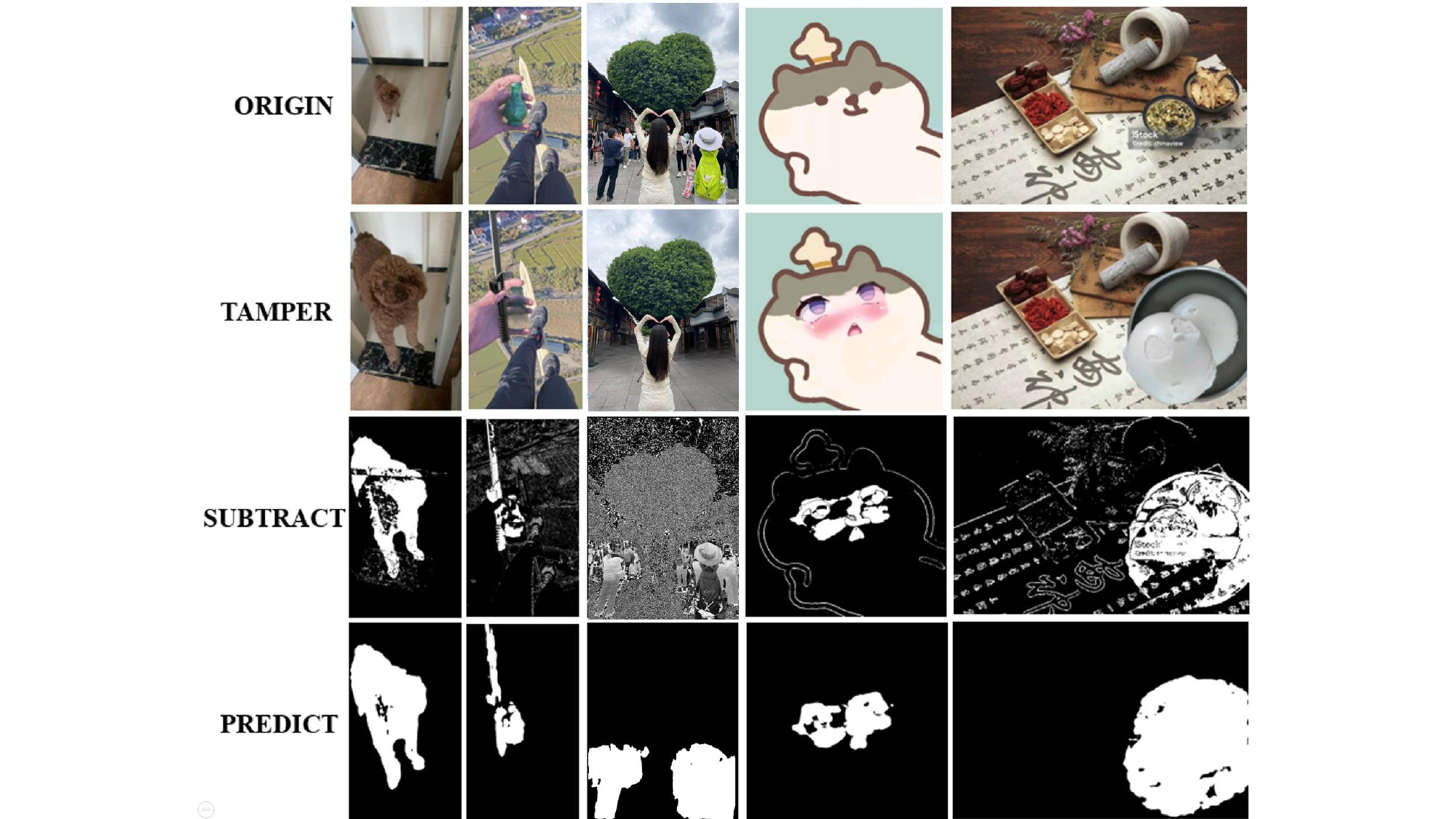}
\caption{MMM framework generated result images. From highest to lowest, the sequence is as follows: original image, tampered image, image obtained by directly subtracting the two images and binarizing with a threshold of 30, and MMM predicted image.} \label{result}
\end{figure*}

However, a common benchmarking image dataset for algorithm evaluation and fair comparison is still lagging behind~\cite{Raise}. Most existing datasets are manually created and annotated by researchers~\cite{MTVB}, with limited tampering types and techniques, and the volume of datasets is also restricted. This leads to models with poor generalization and robustness. Therefore, we thought of creating datasets by sourcing a large number of original and manipulated images from the internet. But we found that images and videos from the internet suffer from compression and clarity issues~\cite{Compression_internet}, and simply subtracting the original and manipulated images results in noisy images, as shown in the third row of Fig. \ref{result}, that traditional methods struggle to clean up.

Change detection~\cite{change_detection1,change_detection2} involves identifying differences at the same location over different times, which is similar to our task of detecting differences between the original and manipulated images. Inspired by this, we treat the original and manipulated images as changes over time for the same picture, viewing the dataset generation task as a change detection task. However, due to the clarity disparity~\cite{crossnet++} between the two images in our task, directly using change detection models is not ideal. Therefore, we introduced a super-resolution processing module to enhance the details of the two images before generating the mask. This is the main idea behind our proposed Manipulation Mask Manufacturer (MMM) framework. Some of the masks we generated and their corresponding original images and tampered images are shown in Fig.\ref{result}.

The framework pipeline inputs the original and tampered images into a Fully Convolutional Network (FCN)~\cite{FCN} to extract high-level features. These features are aligned using Maximum Mean Discrepancies (MMD)~\cite{MMD1,MMD2}
 and concatenated. They are then processed through a Cross-scale Local Attention Block (CSLAB)~\cite{ASSR} and a Local Frequency Encoding Block (LFEB)~\cite{ASSR} to enhance resolution and detail. Finally, the features are split, decoded, subtracted, and a mask is obtained.

\begin{table*}[!ht]
\centering
\caption{The primary datasets in the field of image manipulation localization. Most of these datasets suffer from issues such as limited quantity, single type of tampering, and inability to grow further, and they are all tampered with by the issuers of the datasets themselves.}
\renewcommand{\arraystretch}{1.2} 
\setlength{\tabcolsep}{10pt} 
\resizebox{0.8\textwidth}{!}{ 
\begin{tabular}{cccccccc}
\toprule
\multirow{2}{*}{Dataset} & \multirow{2}{*}{Tampered Images} & \multicolumn{2}{c}{Image Sources} & \multicolumn{2}{c}{Dataset Growth} & \multicolumn{2}{c}{Type} \\
\cmidrule(lr){3-4} \cmidrule(lr){5-6} \cmidrule(lr){7-8}
& & Self & Others & Scalable & Non-scalable & Traditional & AI-manipulated  \\
\midrule
Columbia & 180 & $\checkmark$ & $-$ & $-$ & $\checkmark$ &$\checkmark$ & $-$ \\
CASIAv1 & 920 & $\checkmark$ & $-$ & $-$ & $\checkmark$ &$\checkmark$ & $-$\\
CASIAv2 & 5,063 & $\checkmark$ & $-$ & $-$ & $\checkmark$ &$\checkmark$ & $-$\\
Coverage & 100 & $\checkmark$ & $-$ & $-$ & $\checkmark$ &$\checkmark$ & $-$\\
NIST16 & 564 & $\checkmark$ & $-$ & $-$ & $\checkmark$ &$\checkmark$ & $-$\\
DEFACTO & 149,587 & $\checkmark$ & $-$ & $-$ & $\checkmark$ &$\checkmark$ & $-$\\
IMD20 & 2,010 & $\checkmark$ & $-$ & $-$ & $\checkmark$ &$\checkmark$ &$\checkmark$\\
MMMD(Ours) & 11,069 & $-$ & $\checkmark$ & $\checkmark$ & $-$ &$\checkmark$ &$\checkmark$ \\
\bottomrule
\end{tabular}
}
\label{dataset}
\end{table*}

Our framework has achieved excellent annotation results on the IMD2020~\cite{imd2020}, NIST16~\cite{nist16}, and CASIAv2~\cite{CASIA} datasets. Additionally, we created the Manipulation Mask Manufacturer Dataset (MMMD), containing 11,069 original images, tampered images, and masks, with potential for continuous growth. The dataset includes various resolutions and manipulation types, such as copy-move~\cite{copy_move1,copy_move2}, splicing~\cite{splicing}, transformation~\cite{image_transformation}, Deepfake~\cite{deepfake}, Image Inpainting~\cite{image_inpainting}, Image Morphing~\cite{image_morphing}, Reconstruction~\cite{image_reconstruction}, and Image Style Transfer~\cite{Image_stlye_transfer}. It features diverse images like cartoons, portraits, landscapes, interiors, food, and accessories. The main parameters of our dataset compared to existing datasets are shown in Table \ref{dataset}. We used MMMD to train and test MVSS-Net~\cite{mvss} and IML-ViT~\cite{IML-ViT}. Other 10 models pre-trained on CASIAv2~\cite{CASIA} struggled to achieve high metrics on our dataset, while models trained on our dataset demonstrated better generalization. This highlights the limitations of existing datasets.

In summary, our contributions are as follows:

\begin{itemize}
    \item We propose a Manipulation Mask Manufacturer (MMM) framework that can accurately annotate the differences between original and tampered images even when there is a significant disparity in their clarity.
    \item We generated a large and diverse Manipulation Mask Manufacturer Dataset (MMMD) using the MMM framework to address the shortage of datasets in the field of image manipulation detection.
    \item Models pre-trained with our MMMD achieved higher F1 scores and demonstrated better generalization. Other pre-trained models struggled to perform well on our dataset, highlighting the limitations of existing tampering detection datasets. Our dataset better reflects real-world tampering scenarios.
\end{itemize}

\section{RELATED WORK}

\subsection{Existing dataset generation methods}
Current methods for generating image manipulation detection datasets include manual manipulation, which involves editing images by hand using tools like Adobe Photoshop~\cite{Adobe_Photoshop}, a time-consuming and expertise-demanding process~\cite{manual_M}. Automatic manipulation employs software tools and scripts to rapidly produce large volumes of data, though these images may look unnatural or have obvious manipulation traces. Image synthesis combines elements from different images to create new visual scenes, enhancing dataset diversity but requiring complex techniques and substantial processing time~\cite{Coverage}. The improved Total Variation Denoising Method~\cite{MTVB} automatically subtracts a tampered image from the original to obtain a noisy mask, which is then denoised, but this method often fails to convert many types of tampered images due to diverse noise and low generalization capability of traditional methods.

\subsection{Existing change detection methods}
In recent years, change detection methods based on deep learning have rapidly developed. Hao Chen et al. proposed the Bitemporal Image Transformer (BIT)~\cite{BIT}, which converts input images into a small number of semantic tokens, uses a transformer encoder to model contextual information within the compact token space-time domain, and then projects the context-rich tokens back into the pixel space through a decoder to enhance the original features. The final change detection results are generated through feature difference images.  ChangeFormer~\cite{Changefomer} uses a transformer-based Siamese network architecture for change detection from a pair of co-registered remote sensing images. This method combines a hierarchical transformer encoder and a multi-layer perceptron (MLP) decoder, effectively extracting multi-scale long-range feature differences. This is similar to image manipulation localization, and multi-scale techniques are also commonly used in decoders~\cite{zhu}. In SNUNet-CD~\cite{SNUNet-CD}, ChangeFormer extracts multi-scale features of bitemporal images through a hierarchical transformer encoder and generates change detection maps by fusing these feature differences using a lightweight MLP decoder. However, since the change detection considers the same image at different times with consistent clarity, it does not need to account for image degradation. Therefore, existing change detection models are not effective for our data generation tasks.

\subsection{Existing tampered datasets}
The development of datasets in the field of image manipulation detection has been relatively slow. Currently, widely recognized and used datasets are still those from four or five years ago, or even from over a decade ago. 

\textbf{Datasets for Traditional Tampering Techniques}
Almost all datasets include traditional tampering methods like splicing~\cite{splicing}, copy-move~\cite{copy_move1,copy_move2}, removal, and various image enhancements to produce “fake” or “forged” images. Columbia~\cite{Columbia} uses cropping and splicing~\cite{splicing}, embedding parts from other images into a single image. CASIAv1~\cite{CASIA} employs Adobe Photoshop~\cite{Adobe_Photoshop} for cutting and pasting, including geometric transformations~\cite{image_transformation} like scaling~\cite{image_scaling} and rotation~\cite{image_rotating}. CASIAv2~\cite{CASIA} adds more post-processing and has a richer variety of images, divided into eight categories: scenes, animals, buildings, people, plants, objects, nature, and textures. COVERAGE~\cite{Coverage} consists of real images taken with an iPhone 6 front camera, processed with Photoshop CS4 using methods like translation, scaling, rotation, free transformation~\cite{image_transformation}, lighting changes, and combinations thereof. NIST~\cite{nist16} uses local pixel modification, compression, noise addition, blurring, and geometric transformations~\cite{image_transformation}. DEFACTO~\cite{defacto}, based on the MSCOCO~\cite{MSCOCO} database, aims to produce semantically meaningful forged images, including splicing~\cite{splicing}, copy-move~\cite{copy_move1,copy_move2}, object removal~\cite{object_removal}, and warping~\cite{waring}.

\textbf{Deep Learning-Based Tampering Datasets}
Modern image tampering techniques have achieved unprecedented realism through artificial intelligence and deep learning, particularly with Generative Adversarial Networks (GANs). These techniques include deepfakes~\cite{deepfake}, which can perform facial replacement, expression synthesis, and generate images of non-existent people, making image and video tampering very realistic~\cite{AITamper}. Deep learning also excels in image restoration and enhancement by denoising, filling in missing parts, and improving resolution, thus making damaged images look new and low-resolution images appear clear and detailed. Tools and frameworks such as TensorFlow, PyTorch, Keras, and OpenCV have greatly simplified the implementation and application of these techniques. In IMD2020~\cite{imd2020}, GANs were used to generate tampered regions of images, and inpainting techniques~\cite{image_inpainting} were employed to fill in missing or damaged parts of images, making them appear natural and coherent. This also presents greater challenges for image manipulation detection. Current models are increasingly in need of diverse data that better reflects real-world scenarios.

\section{PROPOSED METHOD}

\begin{figure*}
\centering
\includegraphics[width=0.9\textwidth]{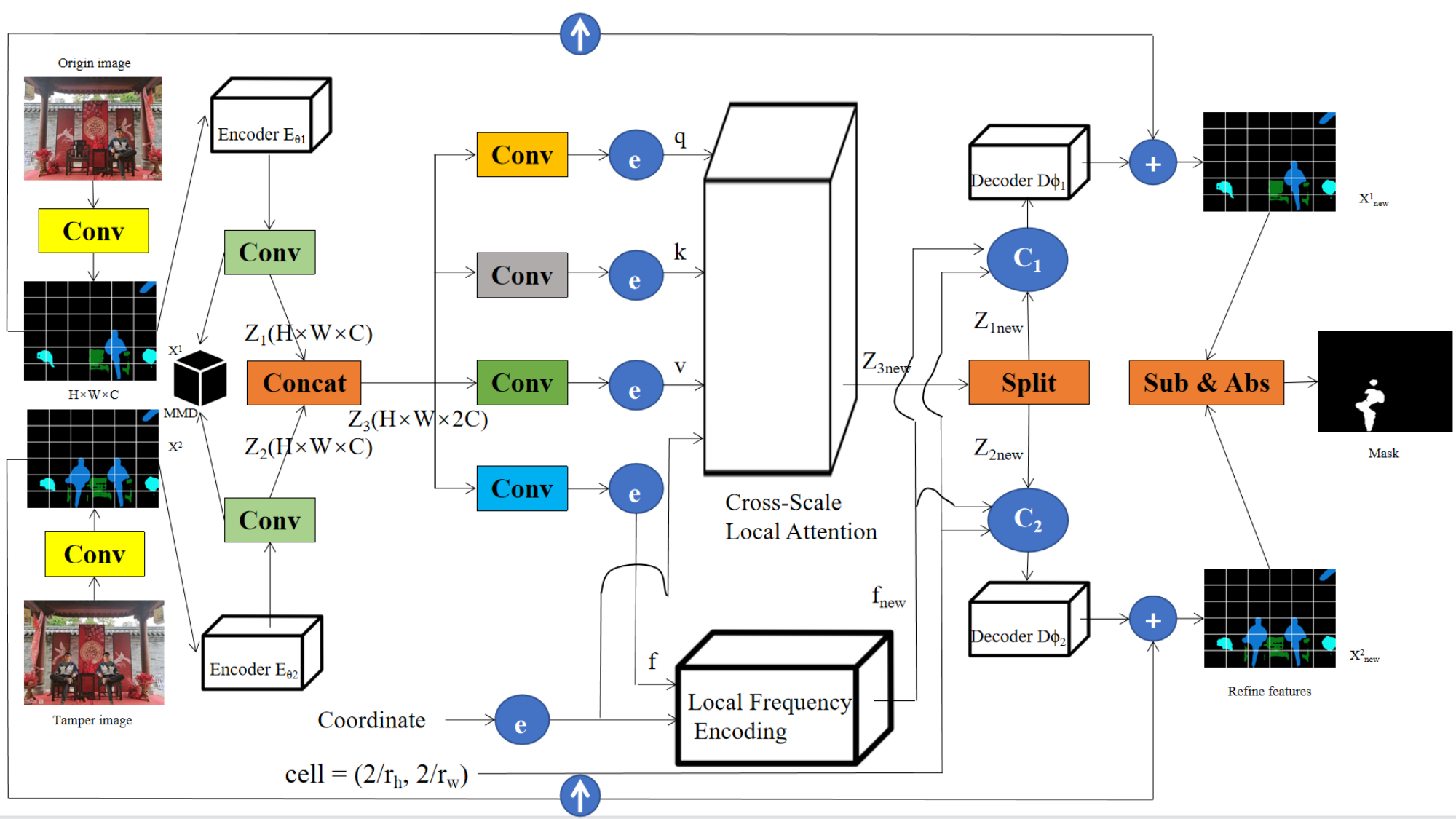}
\caption{The proposed MMM framework. The local sampling operation samples input embeddings based on a grid of coordinates.} 
\label{network}
\end{figure*}

\subsection{The Pipeline of the Entire Framework}
The entire MMM framework is divided into three modules: feature extraction and concatenation, super-resolution processing, and feature separation mask generation. We obtain the original images and a large number of tampered images from the network, keeping only the images of the same size as our original data. After obtaining the original and tampered images, we input them into our Manipulation Mask Manufacturer (MMM) framework. The MMM framework extracts high-level features from the original and tampered images using a Fully Convolutional Network (FCN)~\cite{FCN}. These features are aligned with Maximum Mean Discrepancies (MMD)~\cite{MMD1,MMD2} and concatenated. The idea of concatenating high-level features is inspired by the Bitemporal Image Transformer (BIT)~\cite{BIT}. During super-resolution, the concatenated features are processed by the Cross-scale Local Attention Block (CSLAB)~\cite{ASSR} and the Local Frequency Encoding Block (LFEB)~\cite{ASSR} to enhance the resolution and detail representation of the images. The framework then separates these embeddings, uses decoders to generate residual images, and combines them with the original images. The final mask is produced by computing the absolute difference between the high-resolution features of the original and tampered images. The entire MMM structure is shown in Fig.\ref{network}.


\subsection{Specific Processing Algorithm}

\textbf{Extraction and Concatenation of Image Features}
First, we use a Fully Convolutional Network (FCN)~\cite{FCN} to extract high-level features from the original image and the tampered image, respectively. Then, these features are input into encoder $E_{\theta 1}$ and encoder $E_{\theta 2}$, resulting in the feature embeddings $Z_{1} \in \mathbb{R}^{H \times W \times C}$ and $Z_{2} \in \mathbb{R}^{H \times W \times C}$. $Z_{1}$ and $Z_{2}$ are subjected to Maximum Mean Discrepancies (MMD)~\cite{MMD1,MMD2} calculation to eliminate the differences in data distribution, allowing the model to focus more on the differences in content.  Simultaneously, $Z_{1}$ and $Z_{2}$ are concatenated into $Z_{3} \in \mathbb{R}^{H \times W \times 2C}$.

\textbf{Arbitrary-Scale Super-Resolution}
$Z_{3}$ will be projected by four separate convolutional layers to obtain four latent embeddings, corresponding to query $q$, key $k$, value $v$, and frequency $f$. Since the sizes of the two images are the same, we use the coordinates and cell of the original image. The original image and the tampered image will generate 2D high-resolution coordinates based on an arbitrary upsampling scale $\mathbf{r}=\left\{r_{h}, r_{w}\right\}$ in 2D low-resolution coordinates. Next, the 2D coordinates, along with $q$, $k$, and $v$, will be input into the Cross-Scale Local Attention Block (CSLAB)~\cite{ASSR} to estimate a local latent embedding $Z_{3new} \in \mathbb{R}^{G_{h} G_{w} \times 2C}$. $f$ and the 2D coordinates will also be input into the Local Frequency Encoding Block (LFEB)~\cite{ASSR} to estimate a local frequency embedding ${f_{new}} \in \mathbb{R}^{G_{h} G_{w} \times 2C}$. Specifically, $G_{h}$ and $G_{w}$ represent the height and width of the local grids used for performing local coordinate sampling. CSLAB~\cite{ASSR} and LFEB~\cite{ASSR} estimate $Z_{3new}$ and ${f_{new}}$ as follows:

\begin{equation}
Z_{3new} = CSLAB(\delta x, q, k, v)
\end{equation}
\begin{equation}
f_{new} = \operatorname{LFEB}(\delta x, f)
\end{equation}
\begin{equation}
\delta \mathbf{x} = \left\{ x_{q} - x^{(i,j)} \right\}_{i \in \left\{1, 2, \ldots, G_{h}\right\}, j \in \left\{1, 2, \ldots, G_{w}\right\}}
\end{equation}

\begin{figure}
\centering
\includegraphics[width=0.5\textwidth]{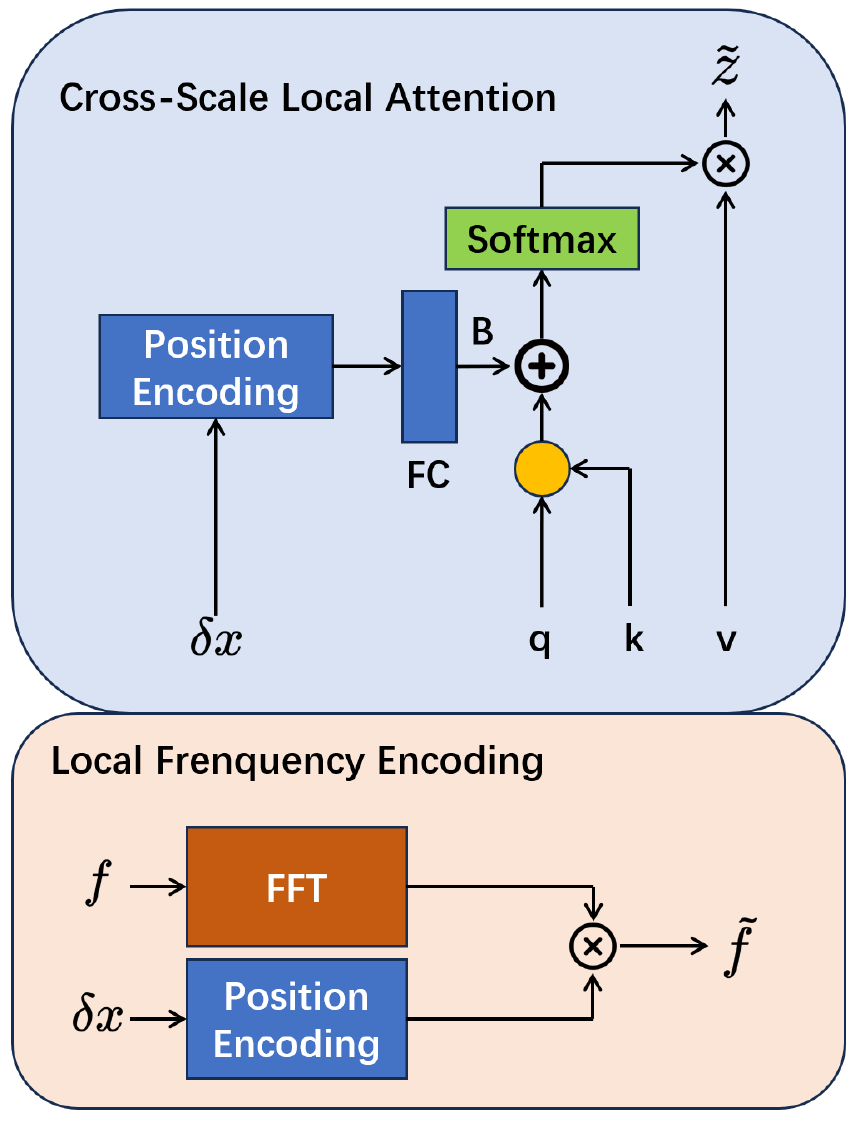}
\caption{The framework of CSLAB and LFEB.}
\label{blocks}
\end{figure}

CSLAB and LFEB draw on the work of Chen et al.~\cite{ASSR}, with specific structures shown in Fig. \ref{blocks}. The primary function of the Cross-scale Local Attention Block (CSLAB) is to aggregate cross-scale local feature information, utilizing query, key, and value features to enhance the model's ability to capture fine-grained details, thereby improving the accuracy of image manipulation detection. The Local Frequency Encoding Block (LFEB) is responsible for performing local frequency encoding on the input frequency features, enhancing the model's sensitivity to edge and texture information by capturing local frequency variations.

\textbf{Separate image features and generate a mask}
$Z_{1new}$ and $Z_{2new}$ are derived from splitting $Z_{3new}$. Decoder $\mathrm{D} \phi_{1}$ and Decoder $\mathrm{D} \phi_{2}$ respectively utilize these embeddings along with the provided cell to generate residual images. Then, the original and tampered images are upsampled and added element-wise to their corresponding residual images. This results in the high-resolution high-level features of the original and tampered images. We subtract these two high-resolution high-level features and take the absolute value to obtain the final mask.

\textbf{Loss Function} The loss function of the model is defined as follows:

\begin{equation}
    L=\frac{1}{H_{p} \times W_{p}} \sum_{h=1}^{H} \sum_{w=1}^{W} l\left(P_{h w}, M_{h w}\right)
\end{equation}

Here, $H_{p} \times W_{p}$ is the total number of pixels in the image, $P_{hw}$ is the probability distribution of the model at $(h, w)$, and $M_{hw}$ is the mask label at position $(h, w)$. $l(P_{hw}, M_{hw})$ is the cross-entropy, which is calculated as:
\begin{equation}
 l(P_{hw}, M_{hw}) = - \sum_{c} M_{hw}^{(c)} \log(P_{hw}^{(c)}) 
\end{equation}
where $c$ indexes the classes, $M_{hw}^{(c)}$ is a binary indicator (0 or 1) if class label $c$ is the correct classification for position $(h, w)$, and $P_{hw}^{(c)}$ is the predicted probability of class $c$ at position $(h, w)$. The cross-entropy measures the dissimilarity between the true label distribution and the predicted probability distribution, and it is used to optimize the model parameters by minimizing this dissimilarity.

\section{EXPERIMENTAL RESULTS AND ANALYSIS}
\subsection{Creation of Manipulation Mask Manufacturer Dataset (MMMD)}
We crawled images from websites containing tampered images on the internet, such as Baidu PS Bar. It is an ideal source for both original and tampered images~\cite{MTVB}. Most users request others to help modify the pictures they provide. As a result, there are often numerous tampered images under their posts. We save the original and tampered images from different posts, allowing us to collect a large amount of data in a short period. We take the first image of each post as the original image, like the first row in Fig. \ref{result}, and consider all other images of the same size in that post as tampered images, like the second row in Fig. \ref{result}.

We then subtract the grayscale images of the original and tampered images to obtain a mask image containing a significant amount of noise. This is because images undergo irreversible compression and quality degradation during transmission over the internet, and they also experience compression when opened with image editing software like Photoshop. The original and tampered images contain noise differences that are imperceptible to the human eye, making the directly subtracted mask image unsuitable for use as training data. The images in the NIST16 dataset are multi-scale, so we used it to pre-train the MMM framework. Therefore, we input the original and corresponding tampered images as pairs into our MMM framework, which consists of three steps: feature extraction and concatenation, super-resolution processing, and feature separation mask generation. This process ultimately produces the predicted mask image shown in the last row of Fig. \ref{result}. The predicted image has significantly reduced noise and can be used for tamper detection models. Since the first image from the PS forum is assumed to be original and subsequent ones tampered, but this isn't always true, it leads to noisy, mostly white masks. Special tampering techniques also cause noisy masks. Therefore, masks with more than 70\% or less than 1\% white area are deemed invalid and removed.

The entire MMMD is divided into three groups: original images, tampered images, and predicted masks, each containing 11,069 images. Each tampered image has a corresponding original image and mask in the other two groups. The dataset contains images with different resolutions and various manipulation types, including copy-move\cite{copy_move1,copy_move2}, transformation~\cite{image_transformation}, Deepfake~\cite{deepfake}, image inpainting~\cite{image_inpainting}, morphing~\cite{image_morphing}, reconstruction~\cite{image_reconstruction}, and style transfer. It encompasses a wide range of image categories, such as cartoons, portraits, landscapes, interiors, food, and accessories.

\subsection{Accuracy of the Model on Existing Datasets}

\begin{table*}[!ht]
\centering
\caption{Performance of Our Model on Different Datasets.}
\renewcommand{\arraystretch}{1.2} 
\setlength{\tabcolsep}{10pt} 
\begin{tabular}{lccccc}
\toprule
\textbf{Dataset} & \textbf{F1} & \textbf{Precision} & \textbf{Recall} & \textbf{IoU} & \textbf{Accuracy} \\
\midrule
IMD2020  & 0.88 & 0.90 & 0.87 & 0.81 & 0.97 \\
NIST16   & 0.94 & 0.95 & 0.92 & 0.89 & 0.98 \\
CASIAv2  & 0.95 & 0.96 & 0.95 & 0.91 & 0.98 \\
\bottomrule
\end{tabular}
\label{D_exp}
\end{table*}
Our innovative use of deep learning for annotating image manipulation detection datasets has no existing comparable methods. Thus, we train and validate our model on the NIST16~\cite{nist16}, IMD2020~\cite{imd2020}, and CASIAV2.0~\cite{CASIA} datasets to demonstrate its effectiveness in distinguishing between original and tampered images. 
The experimental results are shown in Table \ref{D_exp}. The model performs exceptionally well on all three datasets, with high levels across all metrics (F1, Precision, Recall, IoU, and Accuracy). It demonstrates strong generalization capability on image manipulation detection datasets, particularly on datasets involving traditional tampering methods. The performance is slightly lower on the IMD2020~\cite{imd2020}, which uses deep learning techniques such as GANs and inpainting~\cite{image_inpainting}, but overall, the model exhibits good adaptability and performance across various datasets.

\textbf{Implementation Details}
Our models are implemented on PyTorch. Our training is set with a learning rate of 0.01 and a maximum of 100 epochs. The learning rate decay iterations are set to 100. Validation is performed after each training epoch, and the model that performs best on the validation set is used for evaluation on the test set.

\textbf{Evaluation Metrics}
We use the F1-score to evaluate the performance of our model. It balances between tamper detection (recall) and avoiding false positives (precision), preventing the bias that comes from solely pursuing high recall or high precision. The formula for F1-score is as follows:

\begin{equation}
    \text { F1-score }=2 \times\left(\frac{\text { Precision } \times \text { Receall }}{\text { Precision }+ \text { Recall }}\right)
    \label{F1}
\end{equation}

Precision and Recall are expressed using the four metrics: True Positive (TP), False Positive (FP), False Negative (FN), and True Negative (TN). Additionally, we use IoU (Intersection over Union) to represent the model's localization accuracy and Accuracy to evaluate the overall performance of the model. The formula of IoU is $\mathrm{IoU}  =\mathrm{TP}/(\mathrm{TP}+\mathrm{FN}+\mathrm{FP})$ and The formula of Accuracy is $\mathrm{Accuracy} =(\mathrm{TP}+\mathrm{TN}) /(\mathrm{TP}+\mathrm{TN}+\mathrm{FN}+\mathrm{FP})$.

\subsection{Effect of the Generated Dataset on Existing Models}
IMDL-BenCo~\cite{benco} reproduces mainstream IML models, and our model experiments are based on this framework.

\begin{table*}
\centering
\caption{The F1 scores of various models pre-trained on MMMD across different datasets. Left: Pre-trained with CASIAv2, Right: Pre-trained with MMMD (ours).}
\renewcommand{\arraystretch}{1.2} 
\setlength{\tabcolsep}{10pt} 
\resizebox{0.8\textwidth}{!}{ 
\begin{tabular}{lcccc}
\toprule
\diagbox{\textbf{Model}}{\textbf{Dataset}} & \textbf{COVERAGE} & \textbf{Columbia} & \textbf{NIST16} & \textbf{IMD2020} \\
\midrule
MVSS-Net   & 0.26/\textbf{0.34} & 0.39/\textbf{0.49} & 0.25/\textbf{0.28}  & 0.28/\textbf{0.32} \\
IML-ViT  & 0.43/0.29 & 0.78/\textbf{0.81} & 0.33/\textbf{0.34}  & 0.33/\textbf{0.37} \\
\bottomrule
\end{tabular}
}
\label{V_exp0}
\end{table*}

We used our MMMD to train two major models, MVSS-Net~\cite{mvss}, and IML-ViT~\cite{IML-ViT}, and validated their generalization. The results are shown in Table \ref{V_exp0}.

As shown in the table, MVSS-Net and IML-ViT trained using MMMD achieved higher metrics across various datasets compared to models pre-trained on CASIAv2. They exhibit higher generalization and robustness on MMMD. Since MMMD is much larger than commonly used datasets and covers a wider variety of manipulation types and scenarios, it helps improve the generalization ability of existing manipulation detection models. Models pre-trained on MMMD are expected to achieve better performance on other datasets, which is consistent with our experimental results.

\begin{table*}[!ht]
\centering
\caption{The F1 scores of various models pre-trained on CASIAv2 across different datasets.}
\renewcommand{\arraystretch}{1.2} 
\setlength{\tabcolsep}{10pt} 
\resizebox{0.8\textwidth}{!}{ 
\begin{tabular}{lcccc}
\toprule
\diagbox{\textbf{Model}}{\textbf{Dataset}} & \textbf{COVERAGE} & \textbf{Columbia} & \textbf{CASIAv1} & \textbf{MMMD(Ours)}\\
\midrule
Mantra-Net  & 0.08 & 0.46 & 0.12 & \textbf{0.09}  \\
MVSS-Net  & 0.26 & 0.39 & 0.53 & \textbf{0.26}  \\
CAT-Net  & 0.30 & 0.58 & 0.58 & \textbf{0.30}  \\
ObjectFormer  & 0.29 & 0.34 & 0.43 & \textbf{0.32}  \\
NCL-IML  & 0.22 & 0.45 & 0.50 & \textbf{0.26}  \\
TruFor  & 0.42 & 0.86 & 0.72 & \textbf{0.30} \\
IML-ViT  & 0.43 & 0.78 & 0.72 & \textbf{0.24}  \\
PSCC-Net  & 0.23 & 0.60 & 0.38 & \textbf{0.32}  \\
\bottomrule
\end{tabular}
}
\label{V_exp1}
\end{table*} 

We used MMMD to validate image manipulation detection models Mantra-Net~\cite{mantra-net}, MVSS-Net~\cite{mvss}, CAT-Net~\cite{CAT-NET}, ObjectFormer~\cite{objectformer}, NCL-IML~\cite{NCL-IML}, TruFor~\cite{TruFor}, IML-ViT~\cite{IML-ViT} and PSCC-Net~\cite{PSCC}, pre-trained on CASIAv2~\cite{CASIA} and discovered the shortcomings of existing datasets compared to our MMMD. The results are shown in Table \ref{V_exp1}.

As shown in the table, various models pre-trained on CASIAv2~\cite{CASIA} struggled to achieve high metrics on MMMD. CASIAv2 is one of the larger datasets in recent years for manipulation detection, covering a wide range of manipulation types and commonly used as a pre-training dataset for models. However, models pre-trained on CASIAv2 exhibited lower metrics on our MMMD compared to other datasets, highlighting the limitations of traditional datasets. In contrast to our MMMD, these datasets are smaller in size, cover fewer manipulation types, and are still somewhat distant from real-world manipulated images. As a result, models trained on these datasets struggle to perform well on real-life manipulated images.

\textbf{Evaluation Metrics}
We also use the F1-score to measure the accuracy of the model's detection, with the calculation formula given in Equation \ref{F1}.

\textbf{Effect Without Using Super-Resolution}
When processing images directly without using the super-resolution module, for certain severely degraded images, we still obtain noisy images without super-resolution. These images are not suitable for use as training data for manipulation detection models.

\section{CONCLUSION}
In this paper, we creatively propose a Manipulation Mask Manufacturer (MMM) framework for generating image manipulation detection datasets. It addresses the issues of small dataset size, poor quality, and limited types of tampering detection in the field of image manipulation detection. It concatenates image feature embeddings, performs context modeling, and captures long-range relationships between pixels. It uses MMD to eliminate the differences in data distribution. Extensive experiments have validated the effectiveness of our method. We demonstrated the strong performance of the MMM framework on existing datasets. The MMMD dataset we proposed better aligns with the real-world tampering scenarios that manipulation detection models must face. This will help these models improve their generalization and robustness in practical applications. We also demonstrated that MMMD outperforms other datasets in training effectiveness.

\bibliographystyle{plain}

\end{document}